\documentclass[10pt]{article}

\usepackage{supra3d}
\usepackage{amsmath,amsfonts,bm}

\def\eqref#1{equation~\ref{#1}}
\def\1{\bm{1}}

\DeclareMathAlphabet{\mathsfit}{\encodingdefault}{\sfdefault}{m}{sl}
\SetMathAlphabet{\mathsfit}{bold}{\encodingdefault}{\sfdefault}{bx}{n}

\usepackage{hyperref}
\usepackage{url}
\usepackage{booktabs}
\usepackage{array}
\usepackage{graphicx}
\usepackage{xcolor}
\usepackage{amsmath,amssymb}
\usepackage{pifont}
\usepackage{xspace}
\usepackage{placeins}
\usepackage{multirow}
\usepackage{float}
\usepackage{tabularx}
\usepackage{wrapfig}

\DeclareCaptionFont{manuscriptcaption}{\fontsize{9.5}{11.5}\selectfont}
\newcommand{\correspondingmark}{\textsuperscript{\supracorresponding}}
\newcommand{\projectpageurl}{https://supra-3d.github.io/PhysAg3D}

\title{PhysAg3D:}
\suprasubtitle{Bridging 3D Generative and Agentic Models for High-Fidelity and Physically-Interactive Assets}
\suprashorttitle{PhysAg3D}
\hypersetup{
  colorlinks=true,pdfborder={0 0 0},
  pdftitle={PhysAg3D: Bridging 3D Generative and Agentic Models for High-Fidelity and Physically-Interactive Assets},
  pdfauthor={Bi'an Du, Jiemin Fang, Fanpeng Meng, Chen Yang, Pufan Li, Daizong Liu, Qi Tian, Wei Hu}
}
\supraabstract{Generating interactive 3D assets from a single image requires realistic appearance, fidelity to the depicted object, and plausible physical behavior. Yet detailed geometry alone does not determine how an object should move or deform. We introduce PhysAg3D, which couples 3D generative models with agentic reasoning and tool use. A vision-language agent first infers semantic parts, candidate interactions, and physical priors from the image; a 3D generator then produces a textured whole-object mesh from the same image. A hierarchical interaction representation links these revisable priors to mechanisms and the generated surfaces they control, organizing interaction construction and refinement. Guided by this representation and observations of the mesh, the agent uses geometric and physical tools to construct parts, joints, deformable regions, and surface bindings while retaining useful generated geometry and materials. Rendered interactions and execution measurements guide targeted revisions within a fixed budget. The framework accommodates both rigid articulation and compliant deformation. Quantitative and qualitative evaluations on real photographs and external benchmarks demonstrate detailed, image-aligned assets with plausible interactions. Ablations support bounded refinement and source-surface preservation. Project Page: \href{\projectpageurl}{\texttt{supra-3d.github.io/PhysAg3D}}.
}

\author{%
\begin{tabular}[t]{@{}l@{\quad}l@{\quad}l@{\quad}l@{}}
\textbf{Bi'an Du}\textsuperscript{1,2*} &
\textbf{Fanpeng Meng}\textsuperscript{3*} &
\textbf{Jiemin Fang}\textsuperscript{3\ensuremath{\ddagger} \supracorresponding} &
\textbf{Chen Yang}\textsuperscript{3}\\[2pt]
\textbf{Pufan Li}\textsuperscript{1,2} &
\textbf{Daizong Liu}\textsuperscript{4} &
\textbf{Qi Tian}\textsuperscript{2,3 \supracorresponding} &
\textbf{Wei Hu}\textsuperscript{1 \supracorresponding}
\end{tabular}\\[8pt]
\begin{tabular}[t]{@{}l@{\quad}l@{\quad}l@{\quad}l@{}}
\textsuperscript{1}Peking University &
\textsuperscript{2}Guangdong Laboratory of Artificial Intelligence and Digital Economy (SZ) & & \\
\textsuperscript{3}Huawei &
\textsuperscript{4}Wuhan University\\
\textsuperscript{*}Equal Contribution. &
\textsuperscript{\ensuremath{\ddagger}}Project lead. \quad
\correspondingmark Corresponding authors. &
\end{tabular}\\[7pt]
\texttt{\{scncdba, forhuwei\}@pku.edu.cn} \quad
\texttt{\{fpmeng0610,jaminfong\}@gmail.com \quad tian.qi1@huawei.com} \\
}

\begin{document}
\maketitle
\begin{figure}[H]
\centering
\includegraphics[width=\linewidth]{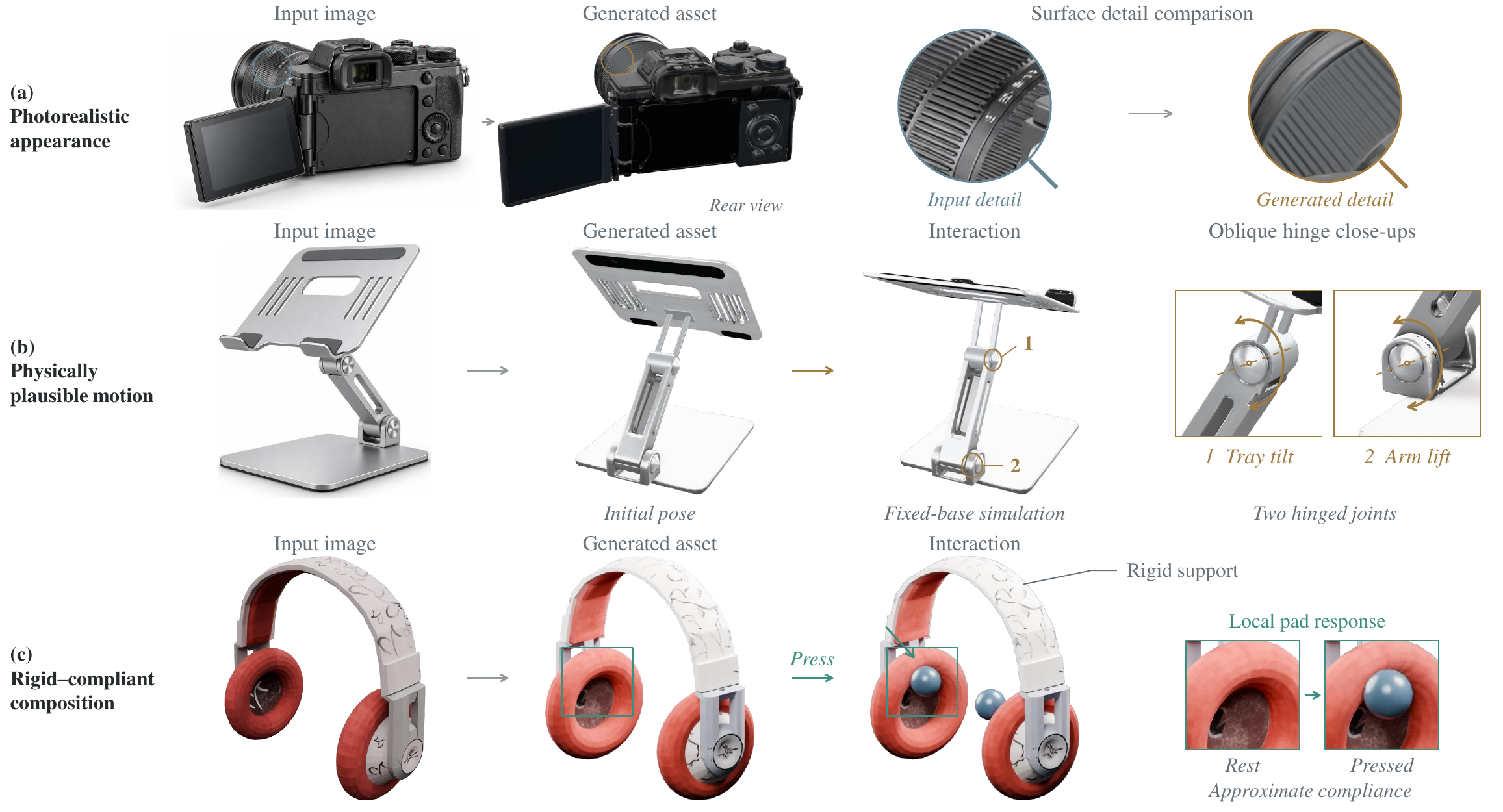}
\caption{\textbf{Detailed appearance with region-specific interaction.} Each example shows an input image, its generated asset, and either appearance details or a selected interaction state. (a) A camera shows detailed surface appearance; insets compare corresponding lens-barrel regions visible in the input image and generated asset. (b) A laptop stand adjusts its tray tilt and support-arm elevation. (c) Headphones combine a rigid support with locally compliant ear pads. These development examples show recorded simulated states in (b) and (c); ear-pad deformation uses an approximate compliance model.}
\label{fig:teaser}
\end{figure}

\clearpage
\flushbottom

\section{Introduction}
\label{sec:introduction}

Generating interactive 3D assets from a single image offers a scalable way to populate virtual environments and embodied simulations with objects from the real world~\citep{mildenhall2020nerf,kerbl2023gaussiansplatting,mo2019partnet,xia2020partnetmobility,xiao2020sapien}. A useful asset should capture the depicted instance, retain realistic geometric and material detail, and respond to actions in ways appropriate to the object. These goals concern different aspects of quality: a visually convincing reconstruction can remain static, while a functional mechanism can belong to an asset that no longer resembles its reference. The task therefore requires bringing appearance, instance alignment, and physical interaction into a coherent realization.

Recent advances provide strong foundations for this task. Image-conditioned generators synthesize detailed geometry and materials~\citep{poole2023dreamfusion,jun2023shape,nichol2022pointe,lin2023magic3d,qian2023magic123,liu2023zero123,shi2023mvdream,liu2023syncdreamer,hong2023lrm,long2024wonder3d,stabilityai2024tripsr, xu2024instantmesh,du2024generative,voleti2024sv3d,wang2024lgm,xiang2024trellis,xiang2025trellis2}; articulated and physical generation methods infer part structure, motion, and material behavior~\citep{mo2019partnet,xia2020partnetmobility,yu2023gapartnet,wang2019shape2motion,xiao2020sapien,liu2025singapo,liu2026pact,cao2026physxomni,makoviychuk2021isaacgym,hu2020difftaichi}; and agentic systems construct and revise assets through executable programs, visual grounding, and feedback~\citep{ahn2022saycan,li2023codeaspolicies,huang2023voxposer,huang2023innermonologue,driess2023palme,brohan2023rt2,xu2023progprompt,valmeekam2023planbench,le2025articulateanything,zhou2026articraft,wang2026adsl}. A remaining challenge is to maintain agreement between what an object is expected to do, how that behavior is implemented, and which visible surfaces actually respond. A single image underdetermines hidden mechanisms and physical parameters, and a generated mesh need not expose the semantic parts required for interaction. Furthermore, a valid joint or deformable simulation does not ensure that the corresponding detailed surface moves with it. Addressing these discrepancies can require changes to geometry and physical structure, creating a further tension between functional construction and preservation of the depicted appearance.

We introduce \textbf{PhysAg3D}, a framework for single-image interactive asset generation through \emph{generative--agentic coupling}. Our central idea is to combine the instance-specific geometry and appearance supplied by a pretrained 3D generator with an agent's ability to reason about object behavior and construct executable physical models. The generated asset provides a detailed visual foundation; agentic reasoning turns this foundation into an interactive realization and revises it in response to observed behavior. This division of capabilities allows physical construction to make use of rich generated surfaces while remaining responsive to the structure and interactions of the depicted object. Figure~\ref{fig:teaser} illustrates this goal through mixer-head articulation, compass-lid closure, telescope reorientation, and local ear-pad deformation. The examples associate actions with specific moving or deforming regions while retaining recognizable object structure and surface detail.

Specifically, a vision-language agent first infers revisable semantic and physical priors from the image, and a 3D generator produces a textured whole-object mesh from the same image. We propose a \emph{hierarchical interaction representation} for this coupling, connecting semantic requirements, physical mechanisms, and concrete asset realizations. Explicit correspondences associate intended interactions with the structures and visible surfaces that implement them. Guided by these relationships and observations of the mesh, the agent uses geometric and physical tools to construct parts, joints or compliant regions, and surface bindings while retaining useful generated geometry and materials. Rendered interactions and execution measurements then guide targeted revisions with bounded construction and validation attempts. The representation expresses both rigid articulation and compliant deformation through their respective relationships between physical state and visible appearance. Our contributions are threefold:
\begin{itemize}[topsep=4pt,itemsep=2pt,parsep=0pt,partopsep=0pt]
    \item A generative--agentic framework that grounds interaction construction in image-derived priors while retaining useful surfaces and materials from detailed whole-object generation.
    \item A hierarchical interaction representation that connects requirements, mechanisms, and concrete surface realizations to organize construction and feedback-guided refinement.
    \item Quantitative and qualitative evaluations on real photographs and external benchmarks, assessing appearance, image alignment, and interaction attributes, with ablations of bounded refinement and source-surface preservation.
\end{itemize}

\section{Related Work}
\label{sec:related_work}

\paragraph{Image-conditioned 3D asset generation.}
Early text-to-3D systems distilled diffusion priors into 3D representations~\citep{poole2023dreamfusion,jun2023shape,nichol2022pointe,lin2023magic3d,qian2023magic123,tang2023makeit3d,du2026part}; feed-forward and multi-view methods improved single-image consistency and speed~\citep{liu2023zero123,shi2023mvdream,liu2023syncdreamer,hong2023lrm,long2024wonder3d,du2024multi,stabilityai2024tripsr,xu2024instantmesh,voleti2024sv3d,wang2024lgm}. TRELLIS introduces structured 3D latents, and TRELLIS.2 supports detailed geometry and physically based materials~\citep{xiang2024trellis,xiang2025trellis2}. UniLat3D unifies geometry and appearance in a shared latent space for single-stage image-to-3D generation~\citep{wu2026unilat3d}. SINGAPO conditions articulated parts on images and part relationships~\citep{liu2025singapo}; PAct jointly generates geometry, composition, and articulation~\citep{liu2026pact}; PhysX-Omni covers rigid, deformable, and articulated objects~\citep{cao2026physxomni}. Shape, surface, and texture reconstruction further improve appearance~\citep{gao2022get3d,li2023neuralangelo,hu2023text2tex,cao2023texfusion,chen2024genesistex,guo2020materialgan,mueller2022instantngp,laine2020nvdiffrast}. More recently, SUPRA-TEX generates relightable materials faithful to reference images by anchoring visible-surface observations in material latent space while completing unobserved regions~\citep{yang2026supratex}. PhysAg3D studies agent-driven construction and revision around a detailed whole-object reconstruction, linking image-derived requirements to mechanisms and visible surfaces.

\paragraph{Articulation and physical modeling of 3D assets.}
PartNet, PartNet-Mobility, GAPartNet, and Shape2Motion support research on parts, affordances, and mobility~\citep{mo2019partnet,xia2020partnetmobility,yu2023gapartnet,wang2019shape2motion}. Articulate AnyMesh segments movable parts and refines joints for open-vocabulary meshes~\citep{qiu2025anymesh}. Particulate predicts segmentation, kinematics, and motion constraints in one pass~\citep{li2025particulate}; Instruct-Particulate adds semantic and point-prompt conditioning, combining 3D generation with vision-language inference for image-to-asset articulation~\citep{li2026instructparticulate}. PhysAg3D focuses on connecting intended interactions, physical entities, exported bindings, and observed surface responses, including compliant regions whose simulation and visible geometry differ.

\paragraph{Agentic 3D modeling and refinement.}
Articulate-Anything uses vision-language reasoning and actor--critic refinement for articulated modeling~\citep{le2025articulateanything}. Language-guided robotics connects instructions to executable actions and closed-loop spatial reasoning~\citep{ahn2022saycan,li2023codeaspolicies,huang2023voxposer,huang2023innermonologue,driess2023palme,brohan2023rt2}, while planning studies emphasize replanning and verification~\citep{xu2023progprompt,valmeekam2023planbench}. ArtiCraft provides a part-and-joint SDK, object-specific tests, and structured execution feedback~\citep{zhou2026articraft}. aDSL unifies hierarchical modeling and spatial relations for planning, execution, and critique, combining procedural scaffolds with TRELLIS through SpaceControl~\citep{wang2026adsl}. Shape programs and CAD languages offer related compositional abstractions and persistent references~\citep{jones2020shapeassembly,jones2025aidl,cascaval2023lineage}. PhysAg3D links requirements, mechanisms, and concrete realizations to generated surfaces. Visual and execution evidence guides local revision while preserving useful geometry.

\paragraph{Simulation and visual grounding substrates.}
MuJoCo, PyBullet, Isaac Gym, and differentiable simulators provide complementary dynamics and contact substrates~\citep{todorov2012mujoco,coumans2016pybullet,makoviychuk2021isaacgym,hu2020difftaichi}. Segment Anything, DINOv2, and CLIP provide visual grouping and image--text alignment primitives~\citep{kirillov2023sam,oquab2023dinov2,radford2021clip}. These form contextual foundations for PhysAg3D's explicit interaction-to-surface representation.

\section{Method}
\label{sec:method}

\subsection{Task Definition and Framework Overview}
\label{sec:overview}

Given a single image $I$, PhysAg3D generates a textured 3D asset with object-specific interactions. An asset contains visible geometry and materials, physical entities and parameters, controls, and bindings from physical state to visible surfaces. We evaluate appearance, image alignment, and physical behavior separately; construction targets the declared interactions, since hidden structure, scale, and material parameters are not determined by one image.

\begin{figure}[t]
\centering
\includegraphics[width=\linewidth]{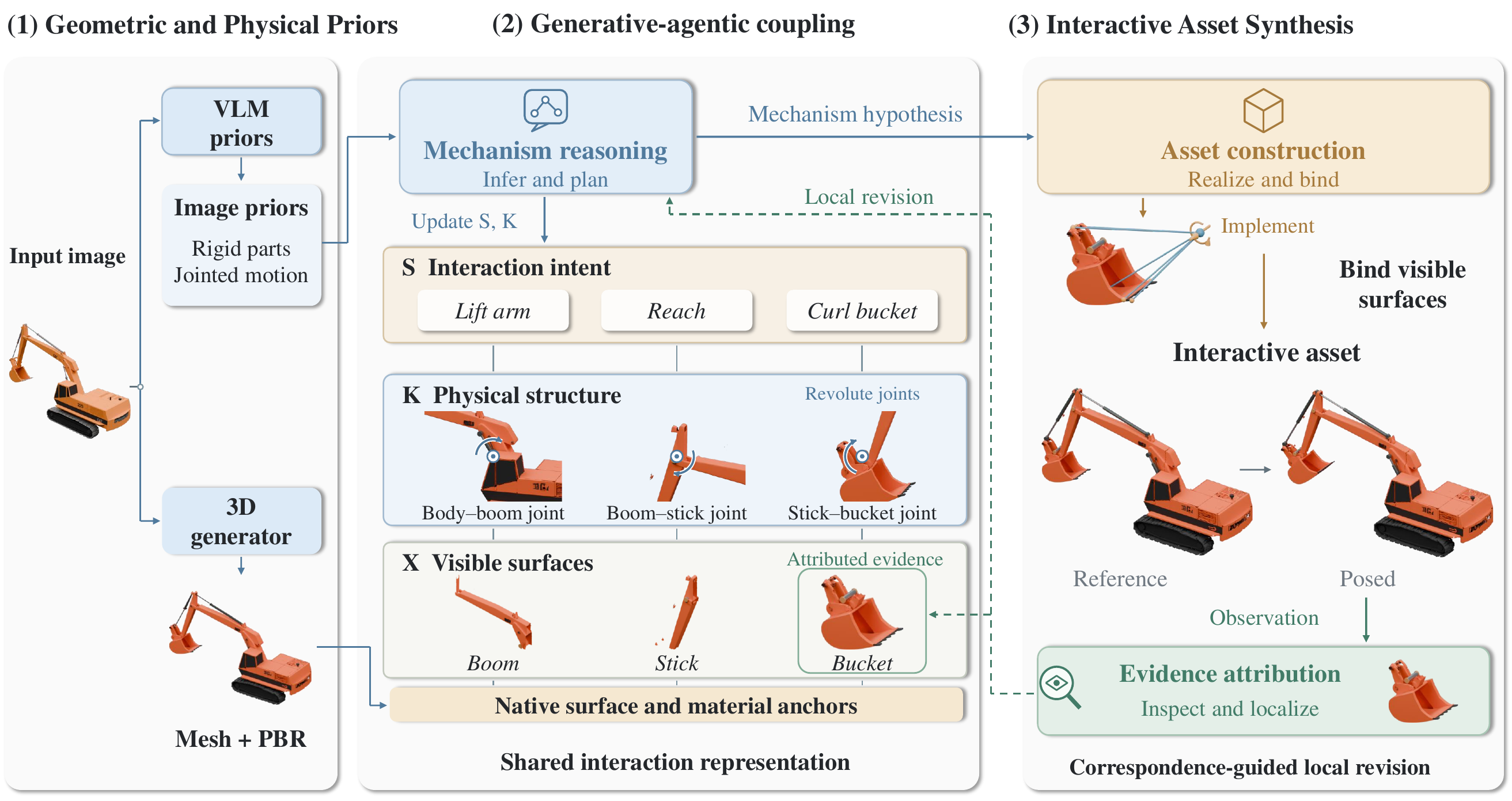}
\caption{\textbf{Overview of PhysAg3D with an excavator example.} (1) Image-derived priors are combined with a generated textured mesh. (2) Interaction intent $S$ specifies lifting, reaching, and curling; physical structure $K$ specifies the corresponding joints; concrete realization $X$ implements them with asset entities and bindings, illustrated here by its visible surfaces. (3) Asset construction binds surfaces to physical entities, and evidence attribution links observations back to the representation for local revision. Mechanism reasoning, asset construction, and evidence attribution operate within one agentic loop. Close-ups show body--boom, boom--stick, and stick--bucket connections; reference and posed states illustrate articulation.}
\label{fig:method}
\end{figure}

Figure~\ref{fig:method} summarizes the workflow. Image interpretation produces revisable semantic and physical hypotheses, while a pretrained generator provides a detailed whole-object mesh. Mechanism reasoning connects these sources through interaction intent $S$ (what should happen), physical structure $K$ (which mechanism produces it), and concrete realization $X$ (the implementing asset entities and bindings). Asset construction realizes these mechanisms and bindings; evidence attribution links observations to the responsible regions for local revision. These are functional roles in one agentic loop. Appendix~\ref{app:implementation} details the interaction representation, construction procedures, validation checks, and candidate-selection policy.

\subsection{Generative--Agentic Coupling}
\label{sec:representation}

\paragraph{Image priors and generated geometry.}
A vision-language agent infers provisional hypotheses $\Pi_0=\Phi(I)$ about parts, materials, connections, candidate interactions, and scale, while retaining ambiguities. A pretrained generator produces a textured whole-object mesh $M_0=G(I)$; our implementation uses TRELLIS.2~\citep{xiang2025trellis2}. The agent inspects rendered views and geometry of $M_0$ before fixing partitions or mechanisms. Priors guide construction but do not condition $G$, so the coupling occurs through interpretation, construction, and feedback without joint training.

\paragraph{Hierarchical interaction representation.}
For candidate version $v$, we define
\begin{equation}
\mathcal R_v=(S_v,K_v,X_v,\Gamma_v,\rho_v),
\qquad \mathcal H_v=(\mathcal R_v,E_v),
\label{eq:representation}
\end{equation}
where $S$, $K$, and $X$ encode semantic requirements, mechanisms, and concrete realizations; $\Gamma$ links their entities; $\rho$ records correspondence to visible regions of $M_0$; and $E$ stores candidate observations. The revision index $v$ is distinct from execution time $t$.

A requirement is
\begin{equation}
 r=(U_r,a_r,\mathcal O_r,\psi_r,\mathcal D_r),
\label{eq:requirement}
\end{equation}
where $U_r$ is the target region, $a_r$ the action protocol, $\mathcal O_r$ the observation, $\psi_r$ the response test, and $\mathcal D_r$ the initial conditions and scope. Thus, lid opening is measured relative to its support, while local compliance is measured after removing common rigid motion. Material compliance and articulation are separate attributes.

The mechanism level $K$ specifies rigid or compliant domains, supports, connections, degrees of freedom, joint frames, limits, and material assumptions. The realization level $X$ identifies actual meshes, materials, scene nodes, controls, simulation entities, and exported bindings, while retaining construction or loading failures. An intended mechanism can therefore remain valid even when its realization is incomplete. In Figure~\ref{fig:method}, ``Interaction intent'' denotes $S$, ``Physical structure'' denotes $K$, and ``Visible surfaces'' depicts the surface component of $X$, whose full scope includes the implementing entities and bindings.

\paragraph{Correspondences across levels and surfaces.}
The relations $\Gamma$ connect requirements to mechanisms, mechanisms to realized entities, semantic parts to visible regions, and physical drivers to bindings. The source map $\rho$ records retained, modified, added, deleted, and unresolved surface/material regions relative to $M_0$. In Figure~\ref{fig:method}, the inter-level links illustrate $\Gamma$, the native surface and material anchors illustrate $\rho$, and attributed observations contribute to $E$. These maps distinguish intended relationships from exported ones: an intended hinged panel may be found attached to the base. Topology changes require an updated correspondence rather than reuse of obsolete face indices. The hierarchy expresses progressively stronger commitments about one candidate, not a fixed sequence of agent roles; requirements can select mechanisms, and observations can retrieve the associated entities for inspection.

\subsection{Physical Construction with Appearance Preservation}
\label{sec:behavior}

The agent grounds semantic parts and compliant domains in image evidence, rendered views, and measurements of $M_0$. Connected geometry can contain multiple moving parts, and texture boundaries need not be mechanical boundaries, so every actionable surface selection is recorded in a declared frame. Construction uses Blender for geometry/material edits and physical assembly tools for simulation entities. Rigid construction sets body ownership, parent--child connections, joint frames and limits, collision geometry, and controls. Compliant construction sets a physical discretization, material assumptions, supports, loads, and a map to the visible skin. Both preserve useful source geometry, UVs, and materials while recording the purpose and source correspondence of bounded edits.

For an executable realization $X$,
\begin{equation}
 z_{t+1}=F_X(z_t,u_t),\qquad
 V_t=B_X(z_t;V_{\mathrm{rest}}),\qquad
 y_{r,t}=\mathcal O_{r,\Gamma,\rho}(V_t,z_t).
\label{eq:behavior}
\end{equation}
Here $F_X$ is the implemented transition, $B_X$ the exported binding, and $y_{r,t}$ the observation used to assess $r$. A kinematic playback and a physics simulator may produce similar visible states, but only the latter supplies dynamics evidence. Rigid vertices follow their assigned body transform; for a compliant region, one valid binding is
\begin{equation}
 V_t=V_{\mathrm{rest}}+W(P_t-P_{\mathrm{rest}}),
\label{eq:soft_binding}
\end{equation}
where $P_t$ are physical-node positions and $W$ maps their displacements to visible vertices in a shared frame. Binding checks verify node identities, dimensions, weights, and rest geometry against the export. A correct rest pose or moving proxy alone does not establish the intended surface response.

\subsection{Refinement with Visual and Physical Feedback}
\label{sec:evidence}
\label{sec:refinement}

Each candidate is evaluated with complementary evidence: renders assess image alignment, retained appearance, part separation, and visible response; source checks assess preserved surfaces and materials; execution checks assess actions, ranges, contacts, supports, and applicable deformation or recovery. MuJoCo~\citep{todorov2012mujoco} is used for supported physical models. A moving proxy, a joint reaching a target, and the intended visible surface responding are distinct observations. Each $E_v$ record stores the requirement, target entities, candidate identity, action, initial state, runtime, and assessment configuration, with outcomes marked unassessed, pass, fail, or inconclusive. Errors without trajectories provide no measured response, and earlier dynamic evidence is invalidated after a build change.

For an unmet requirement, feedback combines the discrepancy or missing evidence with the related mechanisms, realized entities, and bindings obtained through $\Gamma_v$. Given remaining allowances $b_v$, the agent proposes
\begin{equation}
\begin{aligned}
 o_v&\sim\pi(I,\mathcal R_v,E_v,b_v),\qquad
 X_{v+1}=\operatorname{Apply}(X_v,o_v),\\
 \mathcal R_{v+1}&=\operatorname{Reconcile}(\mathcal R_v,o_v,X_{v+1};M_0).
\end{aligned}
\label{eq:revision}
\end{equation}
Correspondences localize inspection but do not guarantee a causal diagnosis: failures may arise from partitions, frames, supports, controls, material assumptions, bindings, or observation definitions. Reconciliation records the actual export and keeps unresolved discrepancies; accepted requirements remain traceable across revisions, so a reduced motion range cannot silently satisfy a full-closure requirement.

A declared policy bounds construction and validation and preserves prior candidates when revisions fail. The system returns the best eligible candidate under a fixed evidence-based selection rule, together with its interactive preview and unresolved requirements. A preview may be useful without a valid simulation trajectory, while its dynamics remain unverified. Appendix~\ref{app:implementation} gives the implementation budgets, checks, and selection policy.
% ，
\section{Experiments}
\label{sec:experiments}
\begin{table*}[tb]
\centering
\caption{Single-image interactive generation on 50 in-the-wild objects with 110 predefined actions. Image metrics use camera-grid-aligned views. ACR, MMR, and RAR are object-level macro averages; FVSP requires all actions to complete without video-visible instability. Art./Def.: articulated/deformable support; $\checkmark$/$\times$: reported or evaluated/no reported support. Best and second-best values are bold and italic, respectively.}
\label{tab:exp-wild}
\scriptsize
\setlength{\tabcolsep}{2pt}
\renewcommand{\arraystretch}{1.15}
\begin{tabularx}{\textwidth}{
@{}l*{9}{>{\centering\arraybackslash}X}@{}
}
\toprule
\multirow[c]{2}{*}{Method}
& \multicolumn{2}{c}{Object Type Support}
& \multicolumn{3}{c}{Input Consistency}
& \multicolumn{2}{c}{Action Quality}
& Range
& \multicolumn{1}{c}{\mbox{Stability}} \\
\cmidrule(lr){2-3}
\cmidrule(lr){4-6}
\cmidrule(lr){7-8}
\cmidrule(lr){9-9}
\cmidrule(l){10-10}
& Art.
& Def.
& FG-IoU$\uparrow$ & PSNR$\uparrow$ & CLIP$\uparrow$
& ACR$\uparrow$ & MMR$\downarrow$
& RAR$\uparrow$
& FVSP$\uparrow$ \\
\midrule

Particulate
& $\checkmark$ & $\times$
& \textit{0.607} & \textit{14.24} & \textit{0.853}
& 46.7\% & 57.5\%
& 34.7\%
& 24.0\% \\

PAct
& $\checkmark$ & $\times$
& 0.531 & 12.53 & 0.806
& 35.0\% & 87.2\%
& 18.0\%
& 16.0\% \\

ArtiCraft
& $\checkmark$ & $\times$
& 0.514 & 12.25 & 0.768
& 56.8\% & 51.8\%
& 34.3\%
& 28.0\% \\

PhysX-Omni
& $\checkmark$ & $\checkmark$
& 0.504 & 13.06 & 0.817
& 41.2\% & 68.5\%
& 26.0\%
& 16.0\% \\

aDSL
& $\checkmark$ & $\times$
& 0.520 & 12.37 & 0.779
& \textit{66.8}\% & \textit{40.2}\%
& \textit{48.8}\%
& \textit{42.0}\% \\
\rowcolor{gray!15}
\textbf{PhysAg3D (ours)}
& $\checkmark$ & $\checkmark$
& \textbf{0.835} & \textbf{16.47} & \textbf{0.892}
& \textbf{91.3\%} & \textbf{22.8\%}
& \textbf{91.3\%}
& \textbf{88.0\%} \\

\bottomrule
\end{tabularx}
\end{table*}

\begin{figure}[tb]
    \centering
    \includegraphics[width=0.95\linewidth]{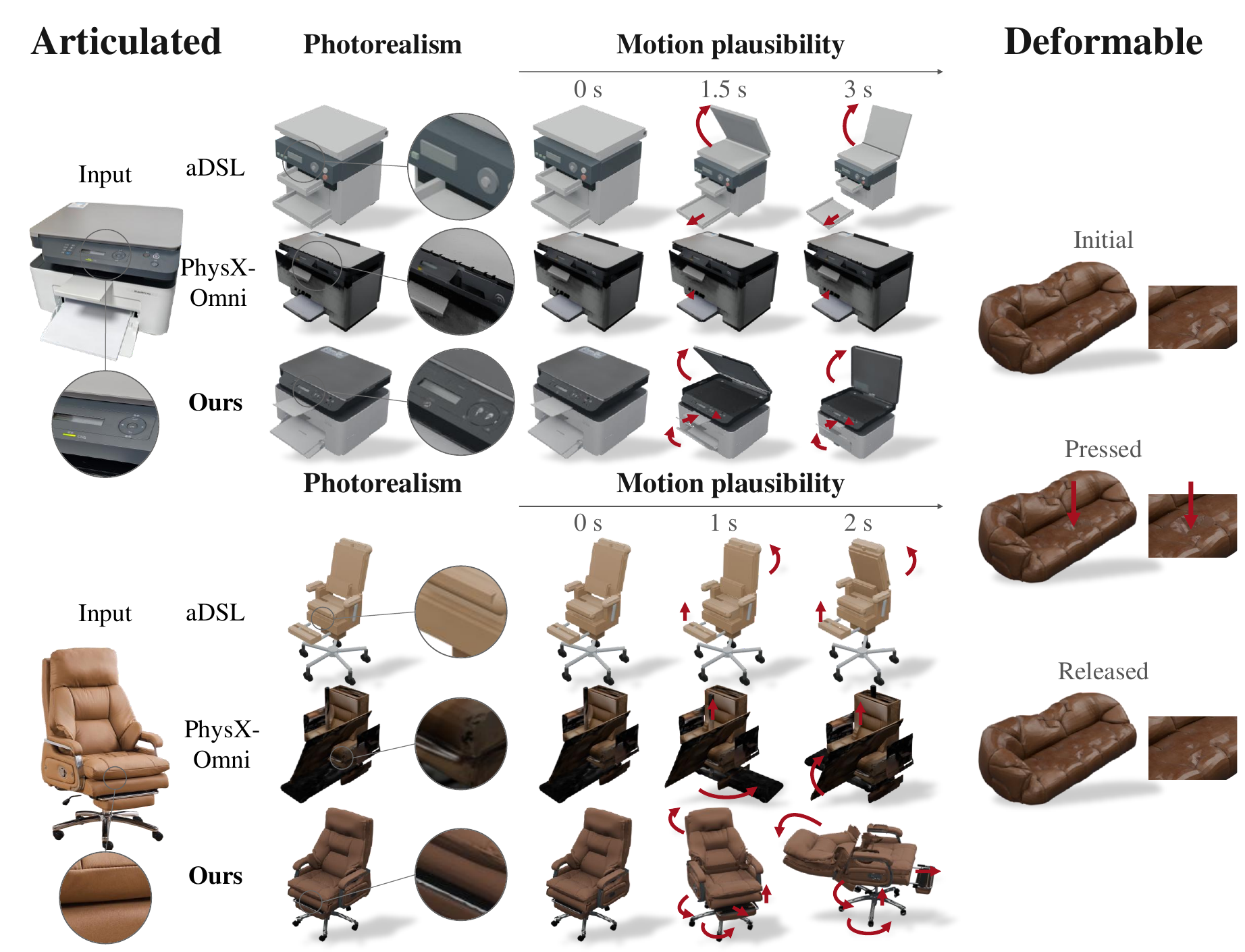}
    \caption{Qualitative comparison on real photographs. Printer and chair panels compare appearance and displayed motion across aDSL, PhysX-Omni, and PhysAg3D. The right column shows initial, pressed, and released states of a deformable asset.}
    \label{fig:exp-wild}
\end{figure}
\subsection{Experimental Settings}
\label{sec:exp-settings}

We evaluate single-image interactive generation on real photographs and external benchmarks, and ablate bounded refinement and source-surface preservation. Implementation and scoring details appear in Appendices~\ref{app:implementation} and~\ref{app:evaluation}.

\paragraph{Common protocol.}
All comparisons fix input lists, method versions, and inference configurations; image-conditioned methods receive the same source image. We record auxiliary conditioning, output support, and computational cost. Candidate selection uses only within-run evidence, without benchmark annotations or final evaluation scores. Refinement permits at most two Builder repairs after initial construction and returns the best available candidate. All attempted inputs remain in the outcome accounting. Static assets are evaluated for input consistency; missing or invalid interaction videos count as failures within the fixed action denominator.

\paragraph{Qualitative development examples.}
Figure~\ref{fig:teaser} shows separate development examples: mixer and compass articulation, approximate compliant ear pads, and telescope poses specified by joint parameters without dynamic validation.

\begin{table}[tb]
\centering
\caption{Reference-based comparison on PhysXVerse and PhysX-Mobility. CD is in $10^{-3}$ and F-score in $10^{-2}$ at threshold $0.05$. Mat., Aff., Kin., and Desc. denote material, affordance, kinematics, and description. Bold and underlined values indicate the best and second-best scores. A dash indicates an entry with no reported score.}
\label{tab:exp-verse}
\scriptsize
\setlength{\tabcolsep}{0pt}
\begin{tabularx}{\linewidth}{@{}l*{8}{>{\centering\arraybackslash}X}@{}}
\toprule
Method & PSNR$\uparrow$ & CD$\downarrow$ & F-score$\uparrow$ & Scale$\downarrow$ & Mat.$\uparrow$ & Aff.$\uparrow$ & Kin.$\uparrow$ & Desc.$\uparrow$ \\
\midrule
\multicolumn{9}{l}{\textit{Dataset: PhysXVerse}} \\
\midrule
Articulate-Anything~\citep{le2025articulateanything} & 14.03 & 48.77 & 46.44 & - & - & - & 0.30 & - \\
MonoArt~\citep{li2026monoart} & 19.68 & 7.03 & 85.27 & - & - & - & 0.38 & - \\
PhysXGen~\citep{cao2025physxgen} & 19.41 & 15.19 & 83.56 & 309.31 & 16.51 & 9.40 & 0.35 & 11.84 \\
PhysX-Anything~\citep{cao2025physxanything} & 15.97 & 37.06 & 40.46 & 298.19 & 15.65 & 10.50 & 0.42 & 21.38 \\
PhysX-Omni~\citep{cao2026physxomni} & \underline{21.52} & \underline{2.95} & \underline{91.28} & \underline{2.79} & \underline{27.23} & 21.47 & 0.92 & \underline{31.05}\\
PAct~\citep{liu2026pact} & 21.10 & 3.20 & 90.90 & 3.05 & 26.50 & 20.80 & 0.92 & 30.30\\
ArtiCraft~\citep{zhou2026articraft} & 16.24 & 12.46 & 77.08 & 5.07 & 20.94 & \underline{27.13} & \underline{0.94} & 23.08\\
aDSL~\citep{wang2026adsl} & 17.18 & 9.76 & 80.93 & 4.16 & 21.87 & 26.14 & 0.93 & 24.92\\

\rowcolor{gray!15}\textbf{PhysAg3D (ours)} & \textbf{24.26} & \textbf{2.08} & \textbf{94.34} & \textbf{2.17} & \textbf{32.18} & \textbf{28.24} & \textbf{0.96} & \textbf{38.16}\\
\midrule
\multicolumn{9}{l}{\textit{Dataset: PhysX-Mobility}} \\
\midrule
Articulate-Anything~\citep{le2025articulateanything} & 15.02 & 16.09 & 66.95 & - & - & - & 0.64 & - \\
MonoArt~\citep{li2026monoart} & 16.46 & 6.35 & 87.41 & - & - & - & 0.44 & - \\
PhysXGen~\citep{cao2025physxgen} & 15.75 & 35.32 & 79.62 & 46.58 & 16.02 & 8.73 & 0.39 & 11.60 \\
PhysX-Anything~\citep{cao2025physxanything} & 16.57 & 23.13 & \underline{89.51} & 22.58 & 22.58 & 16.29 & 0.79 & 26.28\\
PhysX-Omni~\citep{cao2026physxomni} & \underline{18.38} & \underline{4.70} & 88.50 & \underline{2.78} & \underline{24.09} & 16.58 & 0.86 & \underline{28.40}\\
PAct~\citep{liu2026pact} & 18.00 & 5.20 & 86.50 & 3.10 & 23.00 & 16.00 & 0.86 & 27.00\\
ArtiCraft~\citep{zhou2026articraft} & 15.36 & 10.74 & 78.46 & 5.26 & 17.62 & \underline{20.37} & \underline{0.88} & 22.14\\
aDSL~\citep{wang2026adsl} & 16.08 & 8.84 & 81.93 & 4.64 & 19.18 & 19.62 & 0.87 & 23.86\\

\rowcolor{gray!15}\textbf{PhysAg3D (ours)} & \textbf{21.14} & \textbf{2.94} & \textbf{92.56} & \textbf{2.36} & \textbf{29.18} & \textbf{23.16} & \textbf{0.93} & \textbf{34.58}\\
\bottomrule
\end{tabularx}
\end{table}

\begin{table}[tb]
\centering
\caption{PhysX-Bench comparison. RQS: visual quality; MCS: 3D consistency; DQS: absolute scale; APS, KPS, MPS, and DCS denote affordance, kinematic, material, and description scores. All columns are higher-is-better. Bold and underlined values indicate the best and second-best scores, including ties. A dash or NR denotes an unreported score.}
\label{tab:exp-physbench}
\scriptsize
\setlength{\tabcolsep}{0pt}
\begin{tabularx}{\linewidth}{@{}l*{8}{>{\centering\arraybackslash}X}@{}}
\toprule
Method & CLIP & RQS & MCS & DQS & APS & KPS & MPS & DCS \\
\midrule
Articulate-Anything~\citep{le2025articulateanything} & 0.55 & 88.46 & 55.27 & - & - & 71.25 & - & - \\
MonoArt~\citep{li2026monoart} & \textbf{0.84} & \textbf{96.20} & \textbf{82.56} & - & - & 68.32 & - & -\\
PhysXGen~\citep{cao2025physxgen} & \underline{0.80} & 85.93 & 73.50 & 24.21 & 66.07 & 69.17 & NR & 22.24\\
PhysX-Anything~\citep{cao2025physxanything} & 0.55 & 70.81 & 52.71 & 50.20 & 59.96 & 65.99 & 44.70 & 26.89 \\
PhysX-Omni~\citep{cao2026physxomni} & 0.77 & 90.00 & 64.48 & \underline{64.26} & 70.57 & 80.72 & 59.89 & \underline{39.02}\\
PAct~\citep{liu2026pact} & 0.74 & 88.50 & 62.00 & 61.50 & 68.00 & 82.00 & 58.50 & 36.50\\
ArtiCraft~\citep{zhou2026articraft} & 0.49 & 64.18 & 46.12 & 56.24 & 74.16 & \underline{85.88} & 66.24 & 24.18\\
aDSL~\citep{wang2026adsl} & 0.53 & 68.24 & 50.16 & 60.08 & \underline{74.82} & 84.13 & \underline{67.18} & 26.86\\
\rowcolor{gray!15}\textbf{PhysAg3D (ours)} & \textbf{0.84} & \underline{95.18} & \underline{75.24} & \textbf{72.16} & \textbf{82.34} & \textbf{88.18} & \textbf{72.26} & \textbf{48.12}\\
\bottomrule
\end{tabularx}
\end{table}

\begin{figure}[tb]
\centering
\includegraphics[width=\linewidth]{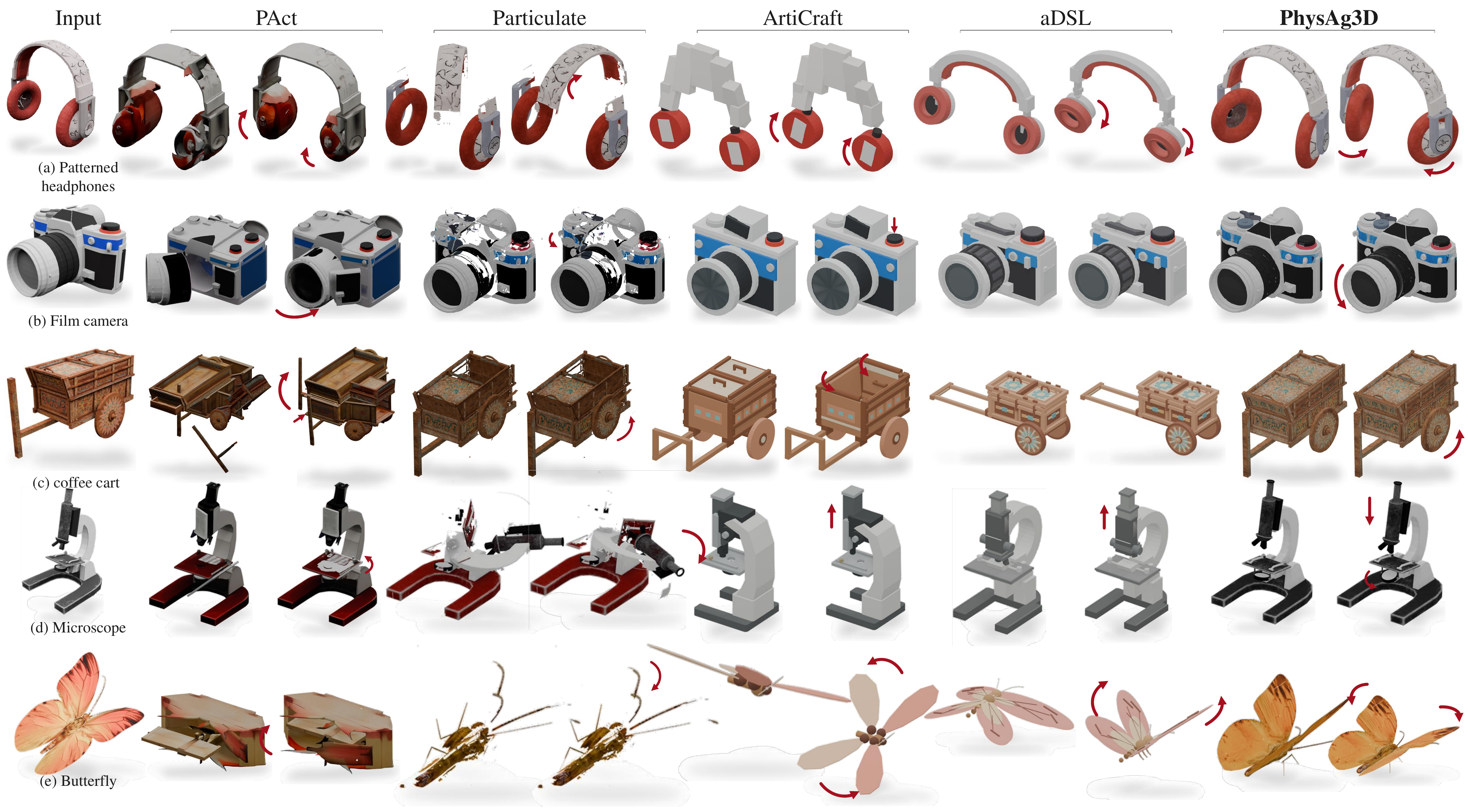}
\caption{\textbf{Qualitative comparison on PhysXVerse.} Each row shows the input and two states per method, with fixed viewpoint and scale within each pair. Red arrows and arcs indicate translation or rotation in our results. The states show appearance and articulation playback, not calibrated dynamics or contact stability.}
\label{fig:exp-physx}
\end{figure}

\subsection{Single-Image Interactive Generation in the Wild}
\label{sec:exp-wild}

\paragraph{Benchmark and comparisons.}
Our benchmark contains more than 200 real-object photographs; evaluation uses a fixed subset of 50 objects with 110 predefined semantic actions, spanning rigid, compliant, and mixed objects. Development cases are identified separately, and method-specific preprocessing is documented. We compare PhysAg3D with Particulate, PhysX-Omni, PAct, ArtiCraft, and aDSL~\citep{li2025particulate,cao2026physxomni,liu2026pact,zhou2026articraft,wang2026adsl}. Additional qualitative comparisons appear in Appendix~\ref{app:qualitative}.

\paragraph{Evaluation metrics.}
For input-view consistency, a fixed spatial grid search selects the rendered camera view closest to the input; foreground IoU, PSNR, and CLIP similarity are computed on the aligned pair. Action Completion Rate (ACR) measures correct semantic completion. Strict Mis-Motion Rate (MMR) counts missing, incomplete, incorrectly related, structurally broken, or spuriously coupled actions. Video-based Range Achievement Rate (RAR) scores the maximum valid visible range in $\{0,0.25,0.5,0.75,1\}$ after the ACR semantic gate. These three metrics are macro-averaged over objects. Function-and-Visible-Stability Pass (FVSP) is the fraction of objects completing all expected actions without video-visible instability. It assesses rendered video evidence, not simulator-log stability.

PhysAg3D leads all reported input-consistency and interaction metrics in Table~\ref{tab:exp-wild}, reaching 91.3\% ACR and 88.0\% FVSP versus 66.8\% and 42.0\% for aDSL, the strongest baseline on these two measures. Figure~\ref{fig:exp-wild} compares appearance and visible interaction states under consistent rendering conditions.

\subsection{Evaluation on External Benchmarks}
\label{sec:exp-physx}

\paragraph{Reference-based evaluation.}
PhysXVerse and PhysX-Mobility provide reference assets for appearance, geometry, and attribute evaluation, with the dataset split reported by PhysX-Omni~\citep{cao2026physxomni}. In Table~\ref{tab:exp-verse}, PhysAg3D achieves the best reported score in every column on both datasets, improving PSNR over the strongest baseline by 2.74 and 2.76 dB, respectively.

\paragraph{Reference-free evaluation.}
PhysX-Bench evaluates generated assets using image and behavior evidence~\citep{cao2026physxomni}. We use its released evidence-generation and scoring pipeline with fixed judge configuration and prompts. In Table~\ref{tab:exp-physbench}, PhysAg3D ties MonoArt on CLIP, ranks second on RQS and MCS, and leads DQS, APS, KPS, MPS, and DCS. These results show strong attribute scores alongside competitive visual quality. The articulation score uses joint playback and does not establish dynamic contact stability.

\paragraph{Qualitative comparison on PhysXVerse.}
Figure~\ref{fig:exp-physx} compares five selected inputs with PAct, Particulate, ArtiCraft, and aDSL~\citep{liu2026pact,li2025particulate,zhou2026articraft,wang2026adsl}. PhysAg3D preserves patterned headphones during earcup rotation, camera housing during lens extension, and ornate cart surfaces during wheel rotation; the microscope and butterfly show localized mechanism and wing motion. ArtiCraft and aDSL simplify fine geometry or patterns, while some PAct and Particulate outputs show fragmented surfaces, misplaced parts, or implausible motion.

\subsection{Human Perceptual Evaluation}
\label{sec:human_eval}

We evaluate six methods for input-image consistency and motion plausibility. First, second, third, and last place receive 5, 4, 3, and 1 points; unranked methods receive 2. First and last place are required, whereas second and third are optional. A separate multi-select question records visible motion collapse.

\begin{table}[t]
    \centering
    \caption{
        Human perceptual evaluation over 102 valid evaluations.
        ``\#1'' denotes first-place selections. ``Collapse'' counts
        evaluations in which visible motion collapse was flagged;
        multiple methods may be selected. Best and second-best results
        are bolded and underlined, respectively.
    }
    \label{tab:user_study}
    \scriptsize
    \setlength{\tabcolsep}{4.5pt}
    \renewcommand{\arraystretch}{1.05}
    \begin{tabular}{@{}lccccc@{}}
        \toprule
        & \multicolumn{2}{c}{Input-image consistency}
        & \multicolumn{2}{c}{Motion plausibility} & \\
        \cmidrule(lr){2-3}\cmidrule(lr){4-5}
        Method
        & Mean$\uparrow$ & \#1$\uparrow$
        & Mean$\uparrow$ & \#1$\uparrow$
        & Collapse$\downarrow$ \\
        \midrule
        Particulate
            & \underline{2.98} & \underline{7}
            & \underline{2.84} & \underline{14} & 21 \\
        PAct
            & 1.99 & 1
            & 1.73 & 0 & 47 \\
        ArtiCraft
            & 2.01 & 1
            & 2.06 & 1 & 28 \\
        PhysX-Omni
            & 2.03 & 5
            & 2.24 & 7 & 15 \\
        aDSL
            & 2.38 & 4
            & 2.75 & 11 & \underline{11} \\
        \textbf{PhysAg3D (ours)}
            & \textbf{4.63} & \textbf{84}
            & \textbf{4.38} & \textbf{69} & \textbf{7} \\
        \bottomrule
    \end{tabular}
\end{table}

Across 102 valid evaluations (Table~\ref{tab:user_study}), PhysAg3D leads in mean input-consistency (4.63 versus 2.98) and motion-plausibility scores (4.38 versus 2.84), and first-place selections. Collapse is flagged 7 times versus 11--47 for baselines. These judgments measure perceived fidelity and motion plausibility, not calibrated dynamics or contact stability.

\subsection{Ablation Study}
\label{sec:exp-ablation}

\begingroup
\setlength{\intextsep}{1pt}
\setlength{\columnsep}{8pt}
\newsavebox{\ablationtablebox}
\sbox{\ablationtablebox}{%
\scriptsize
\setlength{\tabcolsep}{2.5pt}%
\begin{tabular}{@{}lcccc@{}}
\toprule
Setting & CLIP$\uparrow$ & RQS$\uparrow$ & KPS$\uparrow$ & MPS$\uparrow$ \\
\midrule
w/o surface preservation & 0.68 & 77.50 & 75.00 & 59.00 \\
One-shot construction & 0.72 & 82.80 & 68.50 & 55.00 \\
Full pipeline & \textbf{0.84} & \textbf{95.18} & \textbf{88.18} & \textbf{72.26} \\
\bottomrule
\end{tabular}}
\begin{wraptable}{r}{\dimexpr\wd\ablationtablebox+2pt\relax}
\centering
\caption{Ablation on PhysX-Bench.}
\label{tab:exp-ablation}
\usebox{\ablationtablebox}
\end{wraptable}

\paragraph{Ablation settings.}
The three settings share PhysX-Bench inputs, cached TRELLIS.2 reconstructions, image priors, model and decoding settings, construction tools, and final scoring. \textit{Full pipeline} preserves useful source geometry, UVs, and materials while allowing evidence-guided edits and up to two Builder repairs using visual and physical feedback. \textit{One-shot construction} returns the same initial build without repair, but still undergoes final evaluation. \textit{w/o surface preservation} removes the construction-time geometry, UV, and material preservation constraints while retaining the initial reconstruction, physical-to-visible bindings, repair budget, and candidate-selection rule.

\paragraph{Results.}
Table~\ref{tab:exp-ablation} shows that Full pipeline improves KPS and MPS over One-shot by 19.68 and 17.26 points. Removing surface preservation reduces CLIP by 0.16 and RQS by 17.68. These results support bounded refinement and source preservation; failed and inconclusive runs remain in the evaluation.

\FloatBarrier
\section{Conclusion}
\label{sec:conclusion}

We presented PhysAg3D for generating interactive 3D assets from a single image. Its hierarchical representation connects interaction requirements, physical mechanisms, and visible surfaces to guide construction and feedback-based refinement. Experiments support appearance preservation and plausible interactions, with ablations indicating benefits from refinement and surface preservation. Recovering hidden structure and material properties remains a limitation.

%\subsection*{STATEMENT}

%We used generative AI tools for language polishing and assistance with repeated ablation experiment runs and parameter adjustments. The authors manually inspected the execution logs, verified the experimental outputs and reported results, and take responsibility for the final content of this work. 
% The fMRI experiments use the publicly available OpenNeuro ds002748 dataset for secondary analysis; no identifiable patient information or other sensitive personal data are disclosed in this paper or the supplementary materials. An anonymized code package will be provided as supplementary material. Data preprocessing, model configuration, and training procedures are described in Appendix~\ref{append:experimental_details}, while evaluation protocols and ablation settings are documented in Appendix~\ref{sec:additional_ablations}.

\pagebreak[4]
\raggedbottom
% Keep the bibliography compact without changing its text size.
\subsubsection*{Acknowledgments}
This research was financially supported by the Open Research Fund from Guangdong Laboratory of Artificial Intelligence and Digital Economy (SZ), under Grant No. GML-KF-26-29.

\setlength{\bibsep}{3pt plus 1pt minus 1pt}
\bibliography{iclr2027_conference,hierarchy_references}
\bibliographystyle{supra3d}

\clearpage
\appendix
\section*{Supplementary Material}
This supplement provides implementation details, evaluation protocols, and an extended qualitative comparison accompanying the main paper.
\section{Method and Implementation Details}
\label{app:representation}
\label{app:implementation}

\subsection{Interaction Representation}
The representation in Section~\ref{sec:representation} separates interaction intent, mechanism design, and concrete assets. The semantic level $S$ identifies the target region, action, expected response, and observation frame. The mechanism level $K$ specifies bodies or compliant domains, connections, degrees of freedom, supports, and parameter assumptions. The realization level $X$ identifies the meshes, materials, scene entities, controls, and exported bindings. The relations $\Gamma$ connect these levels; $\rho$ records the origin and modification of source surfaces; and $E$ associates observations with the candidate that produced them. These correspondences link interaction requirements to mechanisms, realized entities, and surface bindings for construction and feedback-guided revision, as defined in Section~\ref{sec:representation}. The following subsections detail the associated construction and validation procedures.

For example, a lid-opening requirement refers to the lid region and measures its motion relative to the base. Its mechanism identifies the hinge axis and permitted range, while its realization identifies the moving body and the visible lid faces assigned to it. This correspondence allows the hinge configuration and visible-surface assignment to be inspected separately; it localizes the relevant entities without by itself establishing the cause of a failure. For a padded region, the corresponding records identify the load, support, physical discretization, and visible skin. These examples explain the representation without assuming that hidden mechanisms or material parameters are uniquely recoverable from the input image.

\subsection{Geometry, Materials, and Surface Binding}
\label{app:binding}
The initial TRELLIS.2 reconstruction supplies the detailed visible geometry and materials. Geometric refinement uses rendered views and measurements to identify actionable regions: mesh connectivity alone does not determine which parts should move, and a texture boundary need not be a mechanical boundary. Bounded edits preserve useful source faces, UV coordinates, and material assignments. When topology changes, source correspondence must be updated rather than inherited from obsolete face indices.

For rigid articulation, visible vertices follow the transform of their assigned body. Correct body ownership, local transforms, parent--child relationships, joint frames, and limits are all needed to produce the intended visible motion. Collision geometry can differ from the detailed render mesh, but motion of a collision proxy alone does not verify motion of the visible part.

For compliant regions, the physical discretization and visible skin may also differ. The displacement binding in Eq.~\ref{eq:soft_binding} transfers physical-node displacements to visible vertices in a common frame. Its node identities, dimensions, weights, and rest geometry are checked against the exported asset. Supports and loading conditions are specified separately from the skin binding. A local deformation observation removes common rigid motion before assessing shape change; otherwise, movement of the entire object could be mistaken for local compliance. The ear-pad example uses an approximate compliance model rather than a material calibration recovered from the image.

\subsection{Feedback and Candidate Records}
Rendered observations assess appearance, part separation, and visible response. Execution observations assess the requested action, achieved range, contacts, supports, and applicable deformation or recovery. Each record identifies the requirement, candidate, target region and entities, action, initial state, runtime and assessment configurations, and assessment outcome. The four outcomes are unassessed, pass, fail, and inconclusive. An execution error supplies no measured trajectory, and a changed build requires renewed dynamic evidence.

Feedback associates a discrepancy with the relevant mechanism, realized entity, or binding. Local repairs can address motion range, tracking, surface assignment, or assembly; structural replanning addresses topology, ownership, joint-axis, or material contradictions. Previous candidates remain available when a revision fails. The budgets and selection procedure below bound this loop.

\subsection{Runtime Configuration and Budgets}
Table~\ref{tab:implementation_policy} summarizes the documented v0.2.0 runtime configuration. Blender handles geometry and material editing; MuJoCo executes supported physical models. The reasoning model is accessed through the local Codex interface. The configured generator seed identifies the initialization setting, without implying deterministic agent execution.

\begin{table}[tb]
\centering
\caption{Documented implementation configuration and construction budgets. Host validation attempts and simulations called inside construction tools are counted separately.}
\label{tab:implementation_policy}
\small
\setlength{\tabcolsep}{5pt}
\begin{tabularx}{\linewidth}{@{}lX@{}}
\toprule
Component & Configuration \\
\midrule
3D generator & TRELLIS.2-4B; \texttt{1024\_cascade} route \\
Sampling and texture & 24 sampling steps; 2048-pixel texture size \\
Generator seed & 42 \\
Reasoning model & \texttt{gpt-6-astra}; high reasoning effort \\
Geometry / simulation & Blender / MuJoCo \\
Builder attempts & 3 total: initial construction and at most 2 repairs \\
Host physical validations & At most 2 attempts \\
Structural replanning & At most 1 replan \\
Planner / Reviewer responses & At most 2 / 3 responses \\
Returned role messages & At most 28 \\
Joint-range reductions & At most 2 per joint \\
\bottomrule
\end{tabularx}
\end{table}

Invalid returned messages consume their role-response allowance. Host validation attempts are reserved before execution, so an interrupted simulation also consumes an attempt. Candidate history and cumulative counters persist across continuations; selecting an earlier candidate does not reset them. Simulator calls made inside construction tools are not covered by the host-attempt limit. These counts therefore describe distinct resources rather than an aggregate computation budget. The initial construction and at most two repairs constitute the Builder budget in Section~\ref{sec:exp-settings}; structural replanning does not reset this budget. As in Section~\ref{sec:evidence}, a modified build requires new dynamic evidence, and a preview does not turn an unverified simulation status into a pass.

\subsection{Motion-Range Adaptation and Physical Checks}
For a hinge or slide with original interval $[\ell_0,h_0]$ and neutral coordinate $q_0$, the documented range helper tests
\begin{equation}
\ell_n=q_0+(\ell_0-q_0)(2/3)^n,\qquad
h_n=q_0+(h_0-q_0)(2/3)^n,\quad n\in\{0,1,2\}.
\label{eq:appendix_backoff}
\end{equation}
The original interval remains the reference across repairs. The retained span must be at least $0.2(h_0-\ell_0)$ and, for a hinge, at least $10^{\circ}$. A reduced interval is considered only after a completed failed sweep with a valid rest pose. The checks include rest and both motion directions where applicable. A colliding rest pose, missing clearance information, incomplete sweep, or execution error terminates this search. The achieved range is retained explicitly; reducing travel does not satisfy an unchanged full-closure requirement.

Let $D$ be the positive asset diagonal in meters. The documented acceptance profile uses a contact-proxy penetration tolerance of $0.005D$, an angular endpoint tolerance of $5^{\circ}$, and a translational endpoint tolerance of $0.005D$. Additional checks cover finite states, solver warnings, joint limits, measured interaction, and applicable compliant-region requirements. Geometric sweep checks apply to sampled configurations and do not replace dynamic execution.

For a declared compliant region, a least-squares rigid alignment removes common motion from the simulation nodes. Let $d_{\max}$ be the maximum residual over recorded samples and phase endpoints. The response criterion requires $d_{\max}\ge 0.001D$ together with a pairwise-distance-change check. When recovery is required, the residual after unloading must not exceed $\max(0.001D,0.35d_{\max})$. These criteria are deformation heuristics on simulation nodes; exported-surface response is assessed separately through the binding and visible observations.

\subsection{Output Selection}
Selection excludes unloadable, empty, nonfinite, externally dependent assets and detected source-identity conflicts. In autonomous runs, eligible candidates are ranked by appearance evidence, native retention, semantic evidence, absence of unsupported additions, display checks, declared bodies, supported physics acceptance, and recency, in that order. Native retention includes source identity, face accounting, geometry, and UV/material checks. As specified in Section~\ref{sec:exp-ablation}, Full pipeline and w/o surface preservation use the same selection rule. In the latter, source-geometry, UV, and material preservation constraints are removed from construction, while the shared ranking criteria and physical-to-visible bindings remain. This intervention therefore tests the construction constraints under a fixed output-selection policy. Supported physics acceptance requires a passing report, at least three trajectory states, and matching asset and trajectory identities. A preview can be returned without a valid dynamic trajectory, with that status kept distinct. If no refined candidate is eligible, the native reconstruction can serve as a static fallback. Such a fallback is not evidence that the requested interaction or dynamic validation succeeded; failed and inconclusive attempts remain in the outcome accounting.

\section{Evaluation and Comparison Details}
\label{app:evaluation}

\subsection{Evaluation Scope and Aggregation}
The in-the-wild collection contains more than 200 photographs; the fixed evaluation subset contains 50 objects and 110 predefined semantic actions. Following Section~\ref{sec:exp-wild}, ACR, MMR, and RAR are macro-averaged over objects. For normalized object-level scores $s_i^{(m)}$, metric $m$ is reported as $100N^{-1}\sum_{i=1}^{N}s_i^{(m)}$, with $N=50$. Thus, each object contributes equally, regardless of its number of rendered frames.

ACR measures correct semantic action completion. Strict MMR counts missing, incomplete, incorrectly related, structurally broken, or spuriously coupled actions. RAR scores the maximum valid visible range in $\{0,0.25,0.5,0.75,1\}$ after the ACR semantic gate. FVSP is the percentage of objects that complete every expected action without video-visible instability. It is a joint functional-and-visible-stability criterion based on rendered videos, not simulator logs.
% AUTHOR RECORDS TO CONFIRM: retain the fixed 50-object IDs, subset-selection rule, relation to the full collection, and separation from development cases.
% AUTHOR RECORDS TO CONFIRM: retain per-action ACR/MMR/RAR annotations and range references; specify whether semantic completion and strict motion-defect labels can overlap.

Support marks indicate declared or evaluated capabilities, not successful outputs. Static assets remain evaluable for input consistency; missing or invalid interaction videos count as failures within the fixed action denominator. Failed and inconclusive attempts remain in the outcome accounting, with unsupported output types distinguished in the records. Development examples remain separate from the benchmark comparison.

% \begin{table}[H]
% \centering
% \caption{Evidence and interpretation for the evaluations reported in the main paper. The rows describe evaluation settings, not additional experiments.}
% \label{tab:app_evaluation_details}
% \small
% \setlength{\tabcolsep}{5pt}
% \begin{tabularx}{\linewidth}{@{}p{0.27\linewidth}X@{}}
% \toprule
% Evaluation & Evidence and interpretation \\
% \midrule
% In-the-wild interaction & ACR, MMR, and RAR: macro averages over 50 objects, computed from 110 predefined actions. FVSP: object-level pass rate requiring all actions to complete without video-visible instability. \\
% PhysXVerse / PhysX-Mobility & Reference assets support geometry, appearance, scale, and attribute comparisons. CD uses $10^{-3}$ units; F-score uses $10^{-2}$ units at threshold $0.05$, as in Table~\ref{tab:exp-verse}. \\
% PhysX-Bench & CLIP, RQS, and MCS address alignment, visual quality, and multi-view consistency. DQS addresses scale; APS, KPS, MPS, and DCS address affordance, kinematics, materials, and description. \\
% Component ablation & The same PhysX-Bench setting compares Full pipeline, One-shot, and w/o surface preservation using CLIP, RQS, KPS, and MPS. \\
% Extended visualization & Additional real-world cases and methods support inspection of photorealistic appearance and correspondence to the input image. \\
% \bottomrule
% \end{tabularx}
% \end{table}

\subsection{Inputs, Baselines, and Scoring}
Section~\ref{sec:exp-settings} specifies the common comparison protocol. Image-conditioned methods receive the same photograph; mesh-conditioned methods require an identified reconstruction and any auxiliary conditioning. Format and coordinate conversions are distinguished from geometry repair or inference of missing physical parameters, which would change the evaluated method. PhysXGen denotes the framework introduced in PhysX-3D~\citep{cao2025physxgen}. aDSL reports image-to-shape generation and articulated creation as a downstream application, motivating its inclusion in the single-image comparison.

Reference-based comparisons use compatible alignment, pose, sampling, distance normalization, and scoring settings. Shape normalization must preserve the separately assessed absolute-scale error. Table~\ref{tab:exp-verse} follows the PhysXVerse/PhysX-Mobility split reported by PhysX-Omni; its kinematic score is not a dynamics success rate. Missing scores, marked by a dash or NR, are unreported entries rather than measured zeros.

PhysX-Bench uses fixed evidence generation, evaluator configuration, and prompts across methods. Joint playback supplies articulation evidence, while material response and static appearance are separate observations; playback does not establish dynamic contact stability. The selected states in Figure~\ref{fig:exp-physx} likewise demonstrate visible articulation and surface appearance. Table~\ref{tab:app_evaluation_details} summarizes the evaluation settings.
% AUTHOR RECORDS: supply exact method revisions, evaluator configuration, sample IDs, and baseline score provenance from the experimental records.

\subsection{Ablation Configuration}
The three variants in Section~\ref{sec:exp-ablation} share inputs, cached reconstructions, initial priors, model and decoding settings, construction tools, and final scoring. Full pipeline and w/o surface preservation share the two-repair limit and candidate-selection rule. One-shot returns the initial candidate and receives final scoring without feeding it back for repair. All attempted inputs, including failed and inconclusive runs, remain in the outcome accounting.

The source-preservation ablation removes construction-time geometry, UV, and material constraints while retaining the initial reconstruction and physical-to-visible bindings. It keeps output selection fixed, including the native-retention ranking criterion. The One-shot comparison measures refinement as a whole, including additional repair and candidate-selection opportunities; it does not isolate an individual feedback signal.

\section{Additional Qualitative Comparisons}
\label{app:qualitative}
Figure~\ref{fig:app-wild} extends Figure~\ref{fig:exp-wild} with additional cases and methods, showing geometry, materials, and texture continuity alongside input-image agreement in proportions, parts, colors, and patterns. No additional aggregate score is introduced.
\begin{table}[H]
\centering
\caption{Evidence and interpretation for the evaluations reported in the main paper. The rows describe evaluation settings, not additional experiments.}
\label{tab:app_evaluation_details}
\small
\setlength{\tabcolsep}{5pt}
\begin{tabularx}{\linewidth}{@{}p{0.27\linewidth}X@{}}
\toprule
Evaluation & Evidence and interpretation \\
\midrule
In-the-wild interaction & ACR, MMR, and RAR: macro averages over 50 objects, computed from 110 predefined actions. FVSP: object-level pass rate requiring all actions to complete without video-visible instability. \\
PhysXVerse / PhysX-Mobility & Reference assets support geometry, appearance, scale, and attribute comparisons. CD uses $10^{-3}$ units; F-score uses $10^{-2}$ units at threshold $0.05$, as in Table~\ref{tab:exp-verse}. \\
PhysX-Bench & CLIP, RQS, and MCS address alignment, visual quality, and multi-view consistency. DQS addresses scale; APS, KPS, MPS, and DCS address affordance, kinematics, materials, and description. \\
Component ablation & The same PhysX-Bench setting compares Full pipeline, One-shot, and w/o surface preservation using CLIP, RQS, KPS, and MPS. \\
Extended visualization & Additional real-world cases and methods support inspection of photorealistic appearance and correspondence to the input image. \\
\bottomrule
\end{tabularx}
\end{table}

\begin{figure}[p]
\centering
\includegraphics[width=\linewidth]{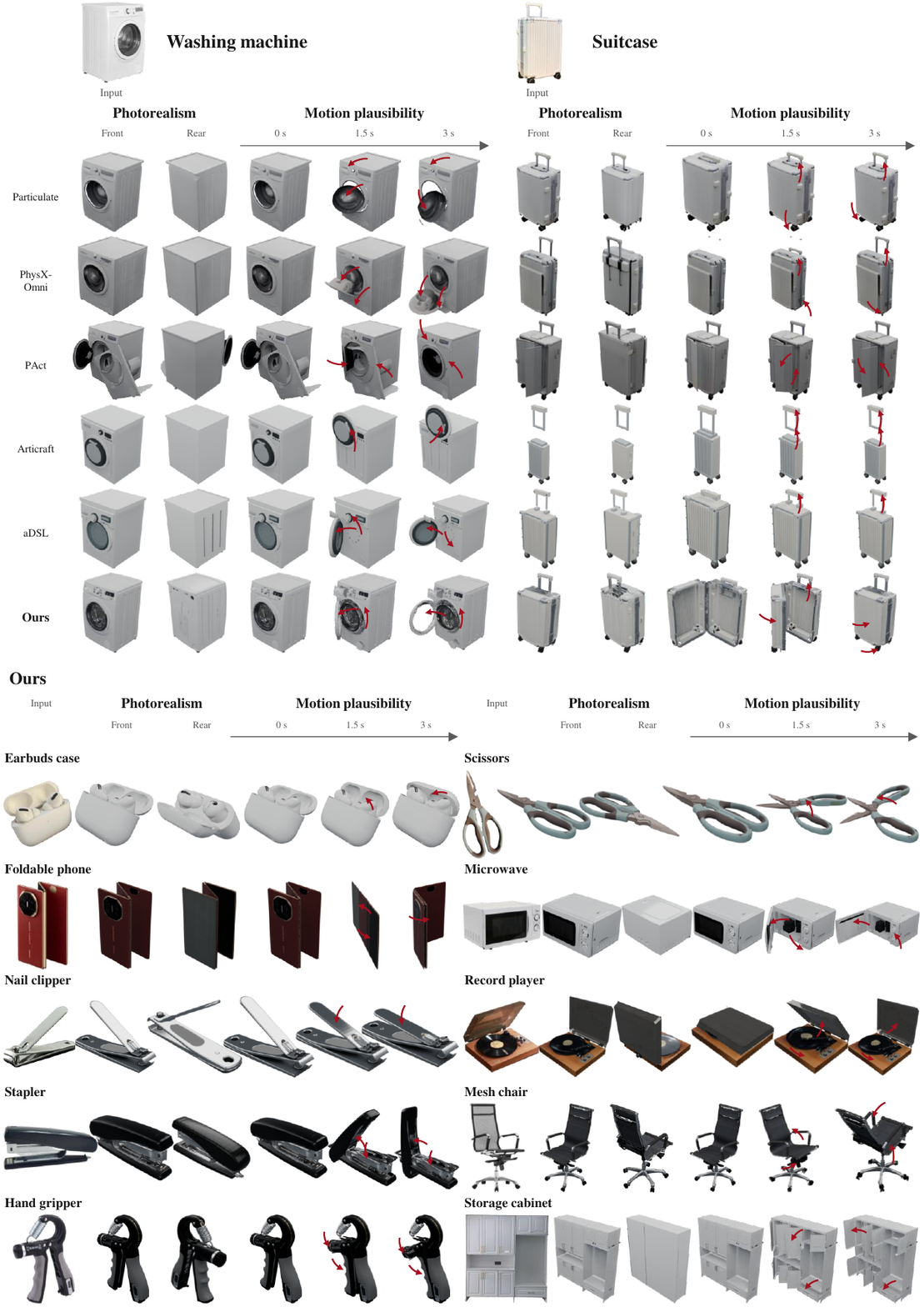}
\caption{\textbf{Extended qualitative comparison on real-world images.} Complementing Figure~\ref{fig:exp-wild}, the expanded comparison focuses on photorealistic appearance and alignment with the input photograph across additional cases and methods.}
\label{fig:app-wild}
\end{figure}

\end{document}